\documentclass{article}

\usepackage{arxiv}

\usepackage[utf8]{inputenc} 
\usepackage[T1]{fontenc}    
\usepackage{hyperref}       
\usepackage{url}            
\usepackage{booktabs}       
\usepackage{amsfonts}       
\usepackage{nicefrac}       
\usepackage{microtype}      
\usepackage{lipsum}
\usepackage{graphicx}
\graphicspath{ {./images/} }

\title{Failure-Aware Refinement of Vision-Language Model for Lithography Defect Detection}

\author{
Pangyun Jeong$^{1,4\dagger}$,
Jiyeong Kong$^{2,4}$,
Yuehua Hu$^{3,4}$,
Dohee Jeong$^{4}$,
Kyung-Tae Kang$^{4*}$\\[0.5em]
$^{1}$HYU-KITECH Joint Department, Hanyang University, Ansan, 15588, Korea\\
$^{2}$Micro/Nano System Department, Korea University, Seoul 02841, Korea\\
$^{3}$Department of Photonics and Nanoelectronics, Hanyang University, Ansan, 15588, Korea\\
$^{4}$Korea Institute of Industrial Technology, Ansan, 15588, Korea\\[0.5em]
$^{\dagger}$First author : \texttt{panguni@hanyang.ac.kr}
$^{*}$Corresponding author: \texttt{kangkt65@gmail.com}
}

\begin{document}
\maketitle

\begin{abstract}
Semiconductor lithography inspection requires reliable detection of small pattern defects such as bridge, burr, pinch, and contamination. In this study, we propose a two-stage vision-language framework that combines initial defect detection with prediction refinement. In the first stage, Qwen3-VL is fine-tuned with LoRA as a vision-language adapter to predict defect counts, defect categories, and normalized bounding boxes from lithography images. However, direct fine-tuning may still produce common test-time errors, including false positives, missed defects, and incorrect defect types. To address this limitation, the second stage trains a refinement module using first-stage prediction failures and their corrected labels, allowing the model to review and revise initial outputs. By learning from cases where the initial adapter fails, the refinement process improves defect inference beyond single-stage fine-tuning.
\end{abstract}


\section{Introduction}
Lithography is one of the most critical processes in modern microfabrication because it defines fine circuit patterns that directly affect device performance, yield, and reliability~\cite{mack2008fundamental}. As pattern dimensions continue to shrink, even small local defects can cause serious manufacturing issues~\cite{levinson2005principles}. Common lithography defects include bridge defects caused by unintended pattern connections, burr defects caused by unwanted protrusions or edge irregularities, pinch defects caused by local line narrowing or disconnection, and contamination defects caused by foreign particles or residues. Detecting these defects accurately is therefore essential for stable process monitoring and quality control~\cite{bunday2014cdsem,orji2018metrology,kim2023deepadc}.

Automated defect inspection has traditionally been addressed using image processing techniques and deep learning-based object detection models~\cite{patel2020defect,dey2022semdefect}. However, these approaches usually require task-specific model architectures, carefully prepared annotation formats, and sufficient labeled data for each defect type. In lithography inspection, this can be challenging because defect shapes are often small, irregular, and visually ambiguous. In addition, an inspection system must often provide not only the presence of defects but also defect categories, counts, and spatial locations in a consistent and interpretable format.

Recent vision-language models have opened a new direction for visual inspection by allowing image understanding and structured text generation within a single framework. Instead of using a fixed detection head, a vision-language model can be fine-tuned to answer inspection-related questions and generate formatted outputs, such as defect counts, class-wise summaries, and bounding box coordinates~\cite{ding2025dietoprompt}. This flexibility makes the model suitable for different structured output requirements~\cite{wang2025yolola}. In this study, we use Qwen3-VL~\cite{bai2025qwen3vl} as a multimodal backbone and fine-tune it with LoRA~\cite{hu2022lora} to generate structured lithography defect predictions from image inputs.

Despite this flexibility, direct fine-tuning of a vision-language model as a single-stage adapter may still produce errors during test. For example, the adapter may detect a non-defective region as a defect, miss a small defect, or assign an incorrect defect type to a detected region~\cite{jiang2024effectiveness}. These failure cases are particularly important in lithography inspection, where small local defects can be difficult to distinguish from normal pattern variations. Therefore, relying only on a single prediction stage may limit the reliability of the final inspection result~\cite{badger2013euvprintability}.

To address this issue, we propose a two-stage vision-language framework that combines initial defect detection with prediction refinement. In the first stage, Qwen3-VL is fine-tuned as a vision-language adapter to predict the total number of defects, defect categories, and normalized bounding boxes~\cite{shen2025vlmr1} from lithography images. The training data are constructed by converting polygon-based annotations into box-formatted instruction-response pairs. In the second stage, a separate refinement module is trained using the first-stage predictions and their corrected labels. This refinement module learns to review the initial output and revise common errors, including false positives, missed defects, and incorrect defect types. By explicitly learning from the failure cases of the initial adapter, the second-stage refinement process aims to produce more reliable defect inference than single-stage fine-tuning.

The main contributions of this study are summarized as follows. First, we formulate lithography defect inspection as a multimodal structured generation task using a vision-language model. Second, we construct a box-based instruction tuning dataset by converting polygon-level lithography defect annotations into normalized bounding-box responses. Third, we propose a two-stage adapter-based refinement framework, where the first-stage adapter generates initial defect predictions and the second-stage refinement module learns to correct failure cases from the initial prediction. Finally, we test the proposed two-stage pipeline on a relatively simple lithography pattern dataset and further test it on more complex pattern images, showing the potential of VLM-based defect inference beyond the training distribution.

\begin{figure}[!htbp]
    \centering
    \includegraphics[width=\linewidth]{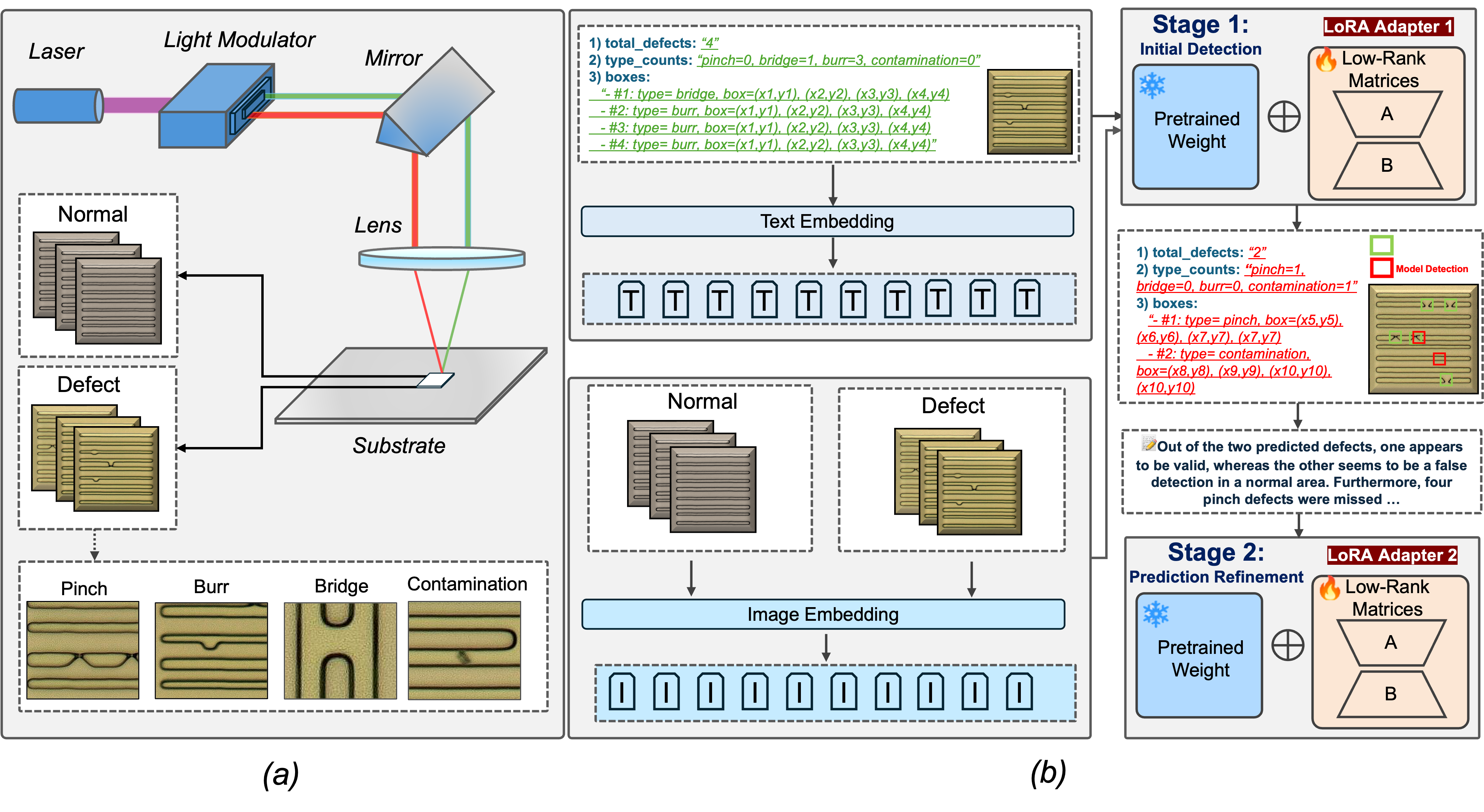}
    \caption{Overall workflow of the proposed two-stage vision-language framework. 
(a) Schematic illustration of lithography defect generation and representative defect classes obtained during the lithography process. 
(b) Proposed two-stage adapter-based refinement pipeline, where the Stage~1 adapter performs initial defect detection and the Stage~2 refinement module reviews and corrects the initial prediction.}
    \label{fig:overall_pipeline}
\end{figure}

The overall workflow of the proposed two-stage framework is illustrated in Figure~\ref{fig:overall_pipeline}. The framework first generates an initial structured defect prediction from the input lithography image and then refines this prediction through a second-stage refinement process.

\section{Dataset}
\label{sec:dataset}

Two lithography datasets are used in this study. The first dataset is used to test the proposed two-stage adapter-based refinement framework, and the second dataset is used to examine its applicability to more complex pattern layouts.

For the main test, we use a lithography defect dataset generated from relatively simple pattern layouts~\cite{hu2025physics}. The dataset contains representative defect types, including bridge, burr, pinch, and contamination. Owing to its clear pattern structures and well-defined defect categories, this dataset is suitable for testing whether the vision-language adapter can learn structured defect inference and whether the refinement stage can improve the initial prediction.

The original pixel- or polygon-level annotations are converted into normalized bounding boxes to match the structured output format of the proposed framework. Each sample is represented as an image-text pair, where the input is a lithography image and the target response includes the total defect count, class-wise defect counts, and bounding box coordinates. The output format is defined as follows:
\begin{quote}
\small
\texttt{1) total\_defects: <int>} \\
\texttt{2) type\_counts: pinch=<int>, bridge=<int>, burr=<int>, contamination=<int>} \\
\texttt{3) boxes:} \\
\texttt{\ \ \ - \#1: type=<class>, box=(x\_min,y\_min,x\_max,y\_max)}
\end{quote}

This dataset is used for both first-stage adapter training and second-stage refinement data construction. For the refinement module, the first-stage predictions are compared with the ground-truth boxes, and correction targets are generated to represent common errors such as false positives, missed defects, and incorrect defect categories.

In addition, we use a more complex lithography pattern dataset derived from the ICCAD-2013 mask optimization benchmark suite~\cite{banerjee2013iccad}. Unlike the simple-pattern test dataset, this dataset contains more complicated layout geometries. In this work, it is used as an additional test set after lithography exposure and microscopic observation. This test is intended to examine whether an adapter trained on relatively simple patterns can still provide meaningful defect inference under more challenging layout conditions.

Overall, the first dataset is used to test the effectiveness of the two-stage adapter-based refinement framework, while the ICCAD-based dataset is used to assess the potential applicability of VLM-based defect inspection to complex lithography patterns.

\section{Method}
\label{sec:method}

The proposed method consists of a two-stage vision-language framework. The first stage performs structured defect detection from lithography images using a vision-language adapter, and the second stage refines the initial prediction by learning from first-stage failure cases.

\subsection{Stage 1: Vision-Language Defect Detection}
\label{subsec:stage1}

Given a lithography image $I$, the first-stage adapter predicts a structured defect output:
\begin{equation}
Y^{(1)} = f_{\theta_1}(I, P_d)
\end{equation}
where $f_{\theta_1}$ is the first-stage vision-language adapter and $P_d$ is the detection prompt. The output $Y^{(1)}$ contains the total number of defects, class-wise defect counts, and a set of predicted defect instances:
\begin{equation}
Y^{(1)} = \{N, \mathbf{c}, \mathcal{B}\}
\end{equation}
Here, $N$ denotes the total defect count, $\mathbf{c}$ denotes the class-wise count vector, and $\mathcal{B}$ denotes the set of predicted defect instances. Each predicted instance is represented as:
\begin{equation}
b_i = (t_i, x_i^{min}, y_i^{min}, x_i^{max}, y_i^{max})
\end{equation}
where $t_i$ denotes the defect class and $(x_i^{min}, y_i^{min}, x_i^{max}, y_i^{max})$ denotes the normalized bounding box coordinates. In this study, $t_i$ belongs to one of four defect classes: bridge, burr, pinch, and contamination.

The training data are constructed by converting polygon-level annotations into bounding boxes. For each polygon, the minimum and maximum vertex coordinates are used to define a normalized bounding box. The resulting image-text pairs are used to fine-tune Qwen3-VL with LoRA. The training loss is computed only on the assistant response tokens, allowing the adapter to learn structured defect prediction conditioned on the input image and prompt.

\subsection{Stage 2: Prediction-Level Refinement}
\label{subsec:stage2}

The first-stage adapter may still produce test-time errors, including false positives, missed defects, and incorrect defect categories. Therefore, a second-stage refinement module is trained to review and correct the initial prediction. The refinement module takes the original image and the first-stage output as input:
\begin{equation}
Y^{(2)} = f_{\theta_2}(I, Y^{(1)}, P_r)
\end{equation}
where $f_{\theta_2}$ is the second-stage refinement module and $P_r$ is the refinement prompt.

To generate refinement training data, the first-stage predictions are compared with the ground-truth boxes using intersection-over-union:
\begin{equation}
\mathrm{IoU}(b_p,b_g) = \frac{|b_p \cap b_g|}{|b_p \cup b_g|}
\end{equation}
Based on the matching result, correction actions are assigned as \texttt{keep}, \texttt{change\_type}, \texttt{remove}, or \texttt{add}. A matched prediction with the correct class is labeled as \texttt{keep}, while a matched prediction with an incorrect class is labeled as \texttt{change\_type}. Unmatched predictions are treated as false positives and labeled as \texttt{remove}, whereas unmatched ground-truth boxes are treated as missed defects and labeled as \texttt{add}.

The refinement module outputs both the correction actions and the final corrected defect result. Since the second-stage refinement module is trained on the failure cases of the first-stage adapter, it learns to revise initial predictions rather than performing detection from scratch. This correction-oriented design improves defect inference beyond single-stage fine-tuning.

\subsection{Inference Pipeline}
\label{subsec:inference_pipeline}

\begin{figure}[!htbp]
    \centering
    \includegraphics[width=\linewidth]{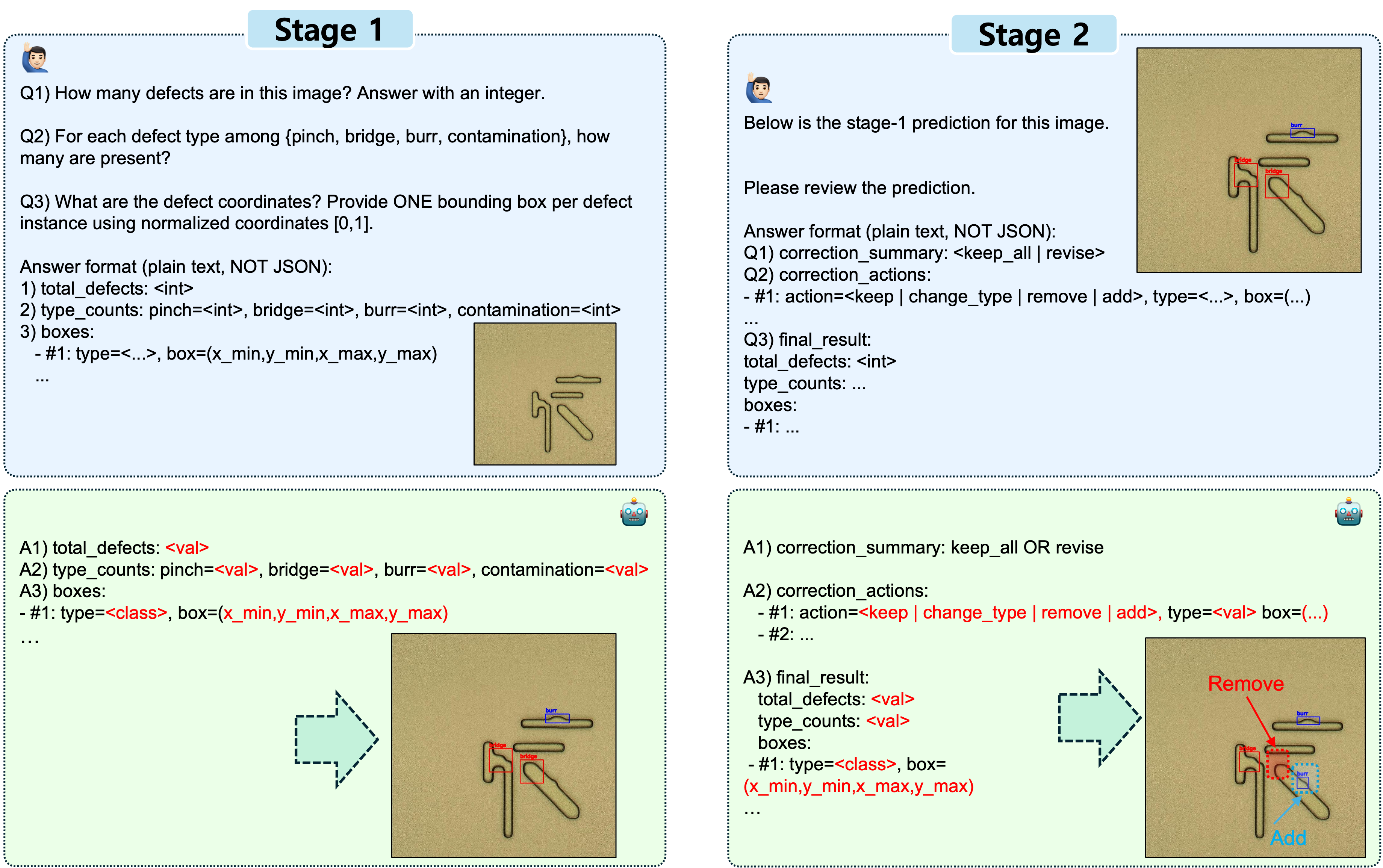}
    \caption{Example of the proposed two-stage inference process. The first-stage adapter generates an initial defect prediction, and the second-stage refinement module reviews the prediction and produces the corrected final result.}
    \label{fig:two_stage}
\end{figure}

At test time, the image is first processed by the first-stage adapter to generate an initial structured prediction. The prediction is then provided to the second-stage refinement module together with the original image. The final output is obtained as:
\begin{equation}
I \rightarrow Y^{(1)} \rightarrow Y^{(2)}
\end{equation}
The final result contains the refined defect count, defect categories, and normalized bounding boxes. Figure~\ref{fig:two_stage} illustrates an example of Stage~1 adapter prediction followed by Stage~2 refinement.

\section{Results}
\label{sec:results}

\subsection{Results on the Simple Defect Pattern Dataset}
\label{subsec:quant_results}

Table~\ref{tab:main_results} presents the quantitative test results of the single-stage VLM adapter and the proposed two-stage adapter-based refinement framework on the simple defect pattern dataset. The single-stage adapter corresponds to direct fine-tuning of Qwen3-VL, while the two-stage framework additionally applies the refinement stage to revise the initial prediction.

Overall, the refinement stage improves AP@0.5 for three out of four defect classes. In particular, the AP@0.5 values for bridge, burr, and contamination increase from 0.805 to 0.843, from 0.740 to 0.799, and from 0.000 to 0.083, respectively. Although the pinch class shows a slight decrease from 0.670 to 0.661, the mAP@0.5 improves from 0.554 to 0.597. These results indicate that the second-stage refinement module is effective in correcting a meaningful portion of the failure cases produced by the first-stage adapter.

\begin{table}[!htbp]
\centering
\caption{AP@0.5 comparison between the single-stage VLM adapter and the proposed two-stage adapter-based refinement framework on the test dataset.}
\label{tab:main_results}
\begin{tabular}{lccccc}
\toprule
Method & Bridge & Burr & Pinch & Contamination & mAP@0.5 \\
\midrule
Single-stage adapter
& 0.805 & 0.740 & \textbf{0.670} & 0.000 & 0.554 \\

Two-stage adapter + refinement
& \textbf{0.843} & \textbf{0.799} & 0.661 & \textbf{0.083} & \textbf{0.597} \\
\bottomrule
\end{tabular}
\end{table}

\subsection{Additional Test on ICCAD-Based Measured Images}
\label{subsec:iccad_results}

To examine the applicability of the proposed pipeline to more complex pattern geometries, we additionally tested the trained adapter on microscope images obtained after actual lithography exposure using ICCAD-based layouts. These images were not used in the training dataset and contain more complex patterns.

\begin{figure}[!htbp]
    \centering
    \includegraphics[width=0.95\linewidth]{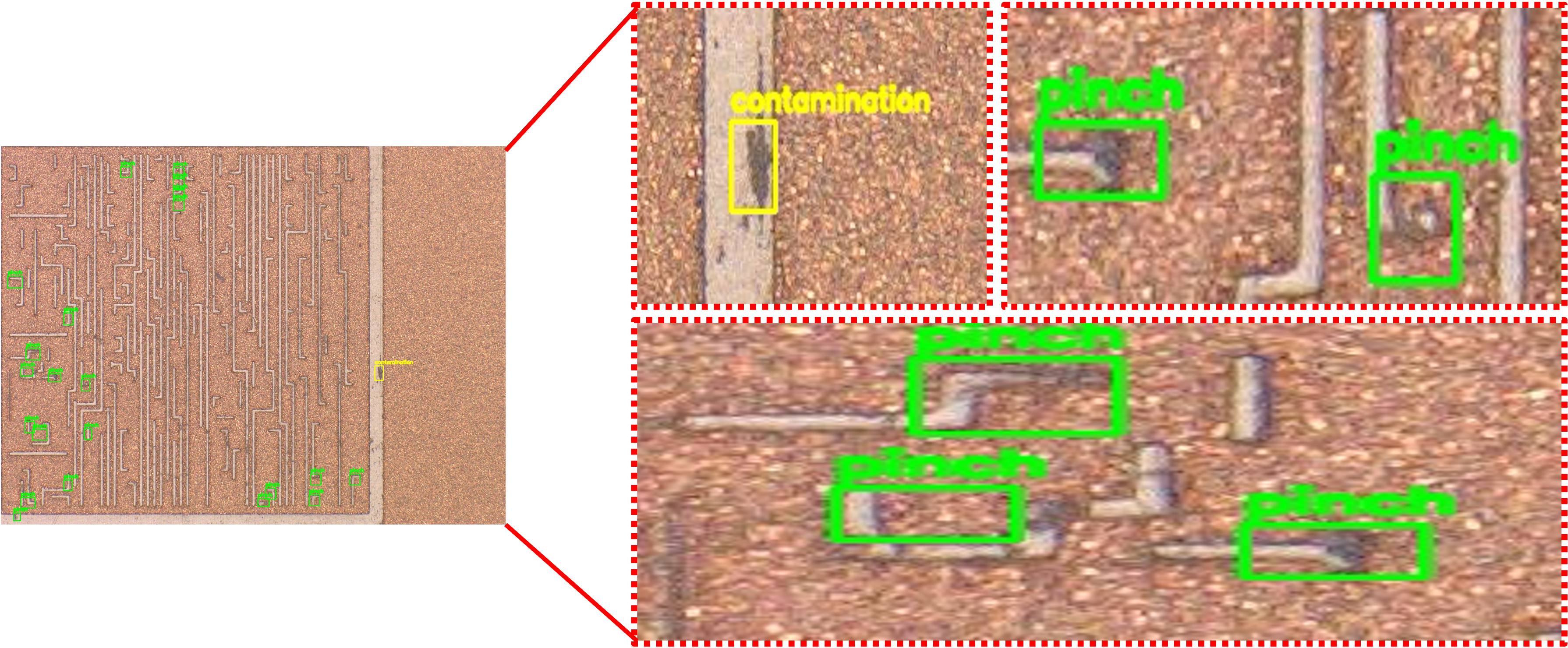}
    \caption{Example inference result on a microscope image obtained after actual lithography exposure for printed wiring board (PWB) using an ICCAD-based layout. The result suggests the potential applicability of the proposed pipeline to more complex measured pattern images.}
    \label{fig:iccad_result}
\end{figure}

Figure~\ref{fig:iccad_result} shows an example result on an ICCAD-based measured image. The pipeline detects several defect candidates, including pinch and contamination, even under more complex layout conditions. This result suggests that a VLM-based adapter trained on relatively simple lithography patterns may still provide useful inference on more challenging measured patterns.

\section{Conclusion}

In this study, we proposed a failure-aware two-stage vision-language framework for lithography defect detection. The first-stage Qwen3-VL adapter generates structured defect predictions, including defect counts, defect classes, and normalized bounding boxes, while the second-stage refinement module reviews and corrects initial prediction errors such as false positives, missed defects, and incorrect defect types.

Experimental results showed that the proposed refinement strategy improved the overall detection performance compared with single-stage fine-tuning, increasing mAP@0.5 from 0.554 to 0.597. In addition, qualitative testing on ICCAD-based measured microscope images suggested that the framework can provide meaningful defect inference even for more complex pattern geometries.

Because lithography is a highly sensitive process, even small undetected defects or incorrect alarms can lead to serious impacts on pattern fidelity, yield, and process reliability. Therefore, AI-based defect inspection should not rely solely on a single prediction stage, but should incorporate multiple verification and correction steps. From this perspective, the proposed two-stage framework provides an industrially meaningful direction for building more reliable and trustworthy defect detection systems in advanced semiconductor manufacturing.

\bibliographystyle{unsrt}  


\end{document}